\documentclass[conference]{IEEEtran}
\usepackage{times}
\usepackage{soul}
\usepackage{url}
\usepackage[hidelinks]{hyperref}
\usepackage[utf8]{inputenc}
\usepackage[small]{caption}
\usepackage{graphicx}
\usepackage{amsmath}
\usepackage{amsthm}
\usepackage{booktabs}
\usepackage{algorithm}
\usepackage{algorithmic}
\usepackage{xspace}
\usepackage{amsmath,amsfonts,bm}

\def\eqref#1{equation~\ref{#1}}
\def\1{\bm{1}}

\DeclareMathAlphabet{\mathsfit}{\encodingdefault}{\sfdefault}{m}{sl}
\SetMathAlphabet{\mathsfit}{bold}{\encodingdefault}{\sfdefault}{bx}{n}

\newtheorem{definition}{Definition}

\usepackage{subcaption}
\usepackage{multirow}
\usepackage{amssymb}
\usepackage{url}

\usepackage{tikz}

\usepackage{verbatim}

\newcommand{\ucan}{UCAN\xspace}

\title{Constrained Non-Affine Alignment of Embeddings}
\author{Yuwei Wang\textsuperscript{\rm 1}, Yan Zheng\textsuperscript{\rm 2}, Yanqing Peng\textsuperscript{\rm 1}, Chin-Chia Michael Yeh\textsuperscript{\rm 2}, Zhongfang Zhuang\textsuperscript{\rm 2}, \\
Das Mahashweta\textsuperscript{\rm 2}, Bendre Mangesh\textsuperscript{\rm 2}, Feifei Li\textsuperscript{\rm 1}, Wei Zhang\textsuperscript{\rm 2}, Jeff M. Phillips\textsuperscript{\rm 1} \\
\textit{\textsuperscript{\rm 1}University of Utah, \textsuperscript{\rm 2}Visa Research} \\
\{ywang, ypeng, lifeifei, jeffp\}@cs.utah.edu\\
\{yazheng, miyeh, zzhuang, mahdas, mbendre, wzhan\}@visa.com}

\begin{document}
\maketitle

%\aistatsauthor{Yuwei Wang\textsuperscript{\rm 1,2} \And Yan Zheng\textsuperscript{\rm 2} \And  Yanqing Peng\textsuperscript{\rm 1} \And Wei Zhang\textsuperscript{\rm 2} \And Feifei Li\textsuperscript{\rm 1}}

%\aistatsaddress{ \textsuperscript{\rm 1}University of Utah, Salt Lake City, USA \And  \textsuperscript{\rm 2}Visa Research, Palo Alto, USA }
%]
%\{ywang, ypeng, lifeifei\}@cs.utah.edu, \{yazheng, wzhan\}@visa.com}

\begin{abstract}
%Merchant embedding is generated by applying natural language processing techniques to the merchant sequence to learn a distributed representations of merchants. As word embedding encodes the syntactic and semantic information, merchant embedding encodes richer information about merchants. In this paper, we introduce a simple but effective way to detect features in the embedding. Using location as an example, we also propose an unsupervised generative adversarial method to align embeddings from different locations together, which is essentially mapping an embedding from one location to another location, while retaining all other properties of the merchants - like price or cuisine type - in their vector representations. We call the procedure embedding location removal. The algorithm is evaluated from three different perspectives, which quantitatively and qualitatively show the effectiveness of proposed methods. 
%Recommendation system has been a hot topic in machine learning fields. There are many start-of-art work on recommendation system based on reviews and so on. However, recommendation system based on transaction data still have more to explore. 

Embeddings are one of the fundamental building blocks for data analysis tasks. Embeddings are already essential tools for large language models and image analysis, and their use is being extended to many other research domains.  The generation of these distributed representations is often a data- and computation-expensive process; yet the holistic analysis and adjustment of them after they have been created is still a developing area.  
In this paper, we first propose a very general quantitatively measure for the presence of features in the embedding data based on if it can be learned.
We then devise a method to remove or alleviate undesired features in the embedding while retaining the essential structure of the data.  We use a Domain Adversarial Network (DAN) to generate a non-affine transformation, but we add constraints to ensure the essential structure of the embedding is preserved. Our empirical results demonstrate that the proposed algorithm significantly outperforms the state-of-art unsupervised algorithm on several data sets, including novel applications from the industry.
\end{abstract}
\begin{IEEEkeywords}
embeddings, alignment
\end{IEEEkeywords}
\section{Introduction}
\label{sec:intro}

An embedding is a moderate-dimensional vector representation of an entity where many features may be captured.  These embeddings are often distributed representations, meaning that in most cases the correspondence between features and vector coordinates is not a one-to-one mapping.  This flexibility allows for better capturing of correlation, the representation of potentially more features than dimensions, and the emergence of a more interesting global structure.  However, it also eschews the interpretability and transparency of analysis of this data, and may result in unwanted associations.  
Yet, due to its efficiency and effectiveness in representing data, embedding learning technology has been an essential part of a wide variety of data domains. In particular, word embedding methods such as Word2vec~\cite{MSC13} have been widely used in natural language processing to capture the semantic and synthetic information about the words. In network management, node embeddings characterize the structures of network nodes~\cite{DBLP:conf/kdd/GroverL16}; while graph embeddings encode rich information of a graph~\cite{cai2018comprehensive}.  Image embedding representations capture essential structural information about individual features~\cite{lowe2004sift} or an aggregate of them~\cite{jia2014caffe}.  
Recent works~\cite{tshitoyan2019unsupervised,grbovic2018real,wang2018billion,zhao2018learning,du2019} also showed that these embedding methods could be applied to even broader ranges of data types. 

%These vectors have been proven to be incredibly useful for capturing similarity between the entities, and have become required components of many downstream learning, mining, and retrieval applications. 
%Encoding structured data as embeddings has three major advantages: 
%(a) it is usually a more efficient representation with lower dimensions than the original data, 
%(b) the nature of the learned representation gives rise to a more expressive representation, and 
%(c) vectors are much easier to use compared to complex structure of the original data. 

Such general embedding schemes have been proven to capture data characteristics without knowing the feature values associated with these characteristics. 
%However, in many cases, analysis can still access features of the original data. 
However, it is still a challenge to recognize when and which features are captured in these embeddings, and how to explicitly identify them, or attenuate them if they are undesired.   
In particular, depending on the downstream task, there might exist undesired known features that have a large impact when analyzing the embedding. In this case, we would like to eliminate their influence in order to improve the performance of downstream tasks. 

We provide powerful methods for recognizing and mitigating such implicitly captured and undesired embedding features.  
We summarize our contributions as follows:
\begin{itemize}
    \item We provide a new method to evaluate the presence and significance of a categorical feature captured by an embedding dataset.  Our evaluation is based on how easily a dataset with multiple labels can be classified.  
    \item We then propose 
    \ucan (\underline{U}nsupervised \underline{C}onstrained \underline{A}lignment that is \underline{N}on-affine), 
    an effective methodology for removing or alleviating the impact of such a significant feature in the embedding.  It uses a variant of a Domain Adversarial Network (DAN) to find a non-affine alignment that obfuscates that feature, but also includes constraints to ensure the original meaning and structure of the embedding is retained. Unlike prior work targeting this goal, the data transformation we learn is not restricted to an affine transformation of part of the data, so it is significantly more powerful, yet the added constraint prevents data deprecation.    
    \item We demonstrate the effectiveness of this process on several embedding datasets representing airports, language, and merchants.  We identify nuisance features, and then train a generator to dampen their effect while retaining the core structure of the embedding.  The usefulness is shown on several downstream tasks, outperforming state-of-the-art methods.  
\end{itemize}

\section{Related Work}
\label{sec:related}

%\textbf{Embedding Bias Attenuation and Mapping.}
Several existing works are focusing on transforming existing embeddings for goals such as attenuating bias. It has been observed that there exists certain (for instance, gender-associated) bias inherent in word embeddings~\cite{bolukbasi2016man,caliskan2017semantics,dev2019attenuating,sweeney2020reducing,dev2019measuring}. In these lines of work, they focused on non-dominating features such as gender, ethnicity, age, and sentiment.  In some cases, they describe simple projection-based~\cite{dev2019attenuating} approaches towards removing the impact of these effects.  
%\cite{dev2019attenuating} uses gendered word pairs to find such minor information and successfully measures it, then attenuates the minor features by projecting the word embeddings away from the subspace occupied by gender.
%However, they cannot deal with features that have dominating effect in the embedding.
Another line of work focuses on bilingual and multi-language embedding alignment. Cross-lingual word embeddings are appealing as they can compare the meaning of words across languages and model transfer between languages (e.g., between resource-rich and low-resource languages, by providing a common representation space). 
However, if the word embedding is trained directly on merged documents from multiple languages, words from the same language tend to cluster due to similar language contexts. As a result, the language of a word becomes the dominating feature embedded in the resulting word embedding. To address this issue, several approaches have been proposed to learn bilingual dictionaries mapping from the source to the target space and align them into the same space using lexicon or a sample of lexicon~\cite{MikolovLS13,AmmarMTLDS16,faruqui2014improving}.
In particular,
~\cite{Conneau2018} learns an initial linear mapping in an adversarial way by additionally training a discriminator to differentiate between projected and actual target language embeddings. \cite{ChenC18} extends this line of work to represent words from multiple languages in a single distributional vector space. This line of work also applies a domain adversarial network.
Our work differs from both of these lines in that it is not restricted to affine transformations and can learn more complex structures of embeddings.

%that it relaxes stringent linearity assumptions in the generator and applies the cosine distance loss as a structure preservation component to keep remaining features in the embeddings. 

There are also similar works from fairness perspective~\cite{madras2018learning,edwards2015censoring,xu2018fairgan,zhang2018mitigating}. They aim to generate fair data using adversarial learning by retaining the ability to predict the label while reducing the possibility of predicting the sensitive or protective variables. In these works, the problem setting is different from ours: the dataset comes with labels; thus, the labels are also involved in the data generation process, while our algorithm only deals with the embedding dataset and does not touch the original label. In addition, this line of work has few discussions on embedding datasets. Although \cite{zhang2018mitigating} discusses in their paper on word embedding dataset, their experiments on this are limited to only show one specific example of the analogy task for the word embedding dataset.

%\input{preliminaries}
%\input{emb_generation}
%\vspace{-1mm}
\section{Feature Measurement in Embeddings}
\label{sec:analysis}

%In this section, we propose a simple but general method to measure the information embedded in the learned embedding. This method does not either depend on the embedding generation methods, or on the entity type. 

Embeddings preserve the information of entities by placing similar entities close together in the embedding space, as measured by cosine distance. 
%As a result, features in the entity will be inherently encoded in the embedding and affect the results of vector operations. 
For example, word embedding captures the semantic and syntactic properties of words, and the embeddings of synonyms are close to each other in the embedding space. However, it is hard to tease apart features since all the features are entangled together, and there is no simple mapping between the dimensions and the features.  Certain traits like polarity (e.g. good - bad) are not specifically encoded by some dimension or direction in a Word2vec embedding. However, for words with similar polarity, like `good' and `great', we observe that they have short cosine distances in the embedding space.

We are interested in a scenario where we can access a subset of features in the original data. In this case, we are looking for an effective way to measure how significantly an embedding is affected by a known categorical feature. We first assume that the \emph{feature} is a binary function $F : \mathcal{D} \to \{0,1\}$ for dataset $\mathcal{D}$ is with $0,1$ labels.  %by which we can separate the data into positive ones and negative ones. %Intuitively, for all positive and negative data points in the data space, the more their embeddings stay apart from each other, the more the feature is embedded. 
The significance of a feature indicates how easily we can distinguish the $0$ and $1$ points.
We quantify the significance using a classifier:

\begin{definition}
Consider an embedding generator $E:\mathcal{D}\rightarrow\mathbb{R}^d$ and a balanced feature $F:\mathcal{D}\rightarrow\{0,1\}$. For a family of classifiers $\mathcal{C}$ on the embedding space, and a positive value $\varepsilon$ with the following probability:

\[
\max_{C \in \mathcal{C}} \text{Prob}_{x\in\mathcal{D}}[C(E(x))=F(x)]>50\%+\varepsilon
\]

We say that $E$ embeds $F$ with weight $\varepsilon$.
\end{definition}

The idea behind this definition is straightforward. If an embedding generated by $E$ does not contain any information about feature $F$, the values of $F(x)$ become random labels for $C(E(x))$. 
Therefore, most classifiers should achieve about $50\%$ accuracy in expectation, on balanced data with half of each label.  
Conversely, if the embeddings can be classified on feature $F$ with accuracy significantly above $50\%$, the embedding reflects some information of $F$. The higher accuracy the classifier can achieve, the more information is encoded in the embedding.  
While we typically cannot precisely find the actual maximum accuracy classifier $C \in \mathcal{C}$, we can use the result of a learning algorithm as a proxy.  

This definition can be extended to multi-label features and classifiers. Suppose the number of labels is $M$, if the classification accuracy is above $1/M$, then the embedding $E$ embeds feature $F$. 
%We delay further details on multi-label challenges to Appendix \ref{sec:multi-label}.  
Numerical features can also be binned to category features using predefined thresholds.

%\vspace{-2mm}
\subsection{Imbalanced Datasets and AUC}

In imbalanced datasets, the ``Accuracy Paradox” occurs when we use the accuracy metric to learn the best model. Consider when a feature has $10\%$ of the data with value $0$, and $90\%$ of the data with value $1$. Then a simple classifier sets $C[E(x)] = 1$ for all $x$ and will get the accuracy of $90\%$, so accuracy is not the best metric for imbalanced datasets. Thus, we use average one vs. all AUC as the metric to measure if a feature is embedded for both balanced and imbalanced data. For a feature $F: \mathcal{D}\rightarrow\{0,1,...,M-1\}$, and any classifier $C$, if $\text{AUC}(C(E(x))=F(x)) > \tau$, then we say \emph{$F$ is embedded by $E$ with weight $\tau$}.
We use this measure within this paper.  

%In this paper, we call a feature a \emph{major feature} if the classification AUC weight $\tau$ is above $80\%$, and a \emph{minor feature} if it is between $50\%-80\%$. 

%\subsection{Numerical Features} 

For fields using numerical values, we can set single or multiple thresholds to label values into different categories. Thus, the AUC metric can be applied to numerical features. The choice of the threshold depends on how much granularity of details needs to be preserved.

\section{Feature Attenuation and Retention}% in Embedding}
\label{sec:disentanglement}
Given an undesired binary feature $F$ on dataset $\mathcal{D}$, let $\mathcal{X}$ and $\mathcal{Y}$ be subsets of $\mathcal{D}$ divided by $F$, and $X,Y \subset \mathbb{R}^d$ be the two sets of corresponding embeddings. For instance, consider embeddings of various merchants generated by the similarity of the transaction sequences from their customers \cite{du2019}.  The subset $\mathcal{X}$ can be merchants in New York City (NYC), and $\mathcal{Y}$ can be merchants from Los Angeles (LA).  To give recommendations for a customer from NYC while visiting LA, based on the embeddings, it would be useful to first remove the effect of the location feature $F$ before doing so.  

%Let $p=|X|$ and $q=|Y|$ be the sizes of these two classes.
%There may not be any alignment between specific elements of $\mathcal{X}/X$ to those of $\mathcal{Y}/Y$.
%Given $X$, $Y$ and the information about $F$ (i.e., we know that embeddings in $X$/$Y$ are generated by data points with label $0$/$1$ on $F$),
The goal of a Domain-Adversarial Network (DAN)~\cite{ganin2016domain} is to find a mapping function $G : \mathbb{R}^d \to \mathbb{R}^d$
(which is also our generator) for $X$ without knowing the data of $\mathcal{X}$ and $\mathcal{Y}$,
%not so there is necessarily a direct alignment, but so their distributions in aggregate are indistinguishable.  
so that sets $Y$ and $G(X)$ are indistinguishable.  
If there was a pattern among the features of $X$ distinct from those of $Y$, it could potentially separate them.  So as an added benefit, if an element $x \in \mathcal{X}$ is similar to some $y \in \mathcal{Y}$ after the feature $F$ has been removed, then the resultant embedding $G(x)$ will tend to be similar to that of $y$.   

%if two elements $\mathbf{x}\in \mathcal{X}$ and $\mathbf{y}\in \mathcal{Y}$ are similar without the presence of feature $F$, then the map result $G(x)$ will have a close distance to the embedding of $y$. 
We name embedding $X$ the \emph{source domain} (e.g., NYC merchants) and embedding $Y$ the \emph{target domain} (e.g., LA merchants).  
A discriminator $D : \mathbb{R}^d \to [0,1]$ is trained to distinguish between elements randomly sampled from $G(X) = \{G(x_1), ..., G(x_p)\}$ (value closer to $0$) and $Y = \{y_1, \ldots, y_q\}$ (values closer to $1$). 
Generator $G$ is trained to prevent the discriminator from making accurate predictions. 
As a result, this is a two-player game, where the discriminator aims at maximizing its ability to identify the origin of an embedding, and $G$ aims at preventing the discriminator from doing so by making the distribution of $G(X)$ and $Y$ as similar as possible on $F$. After the network converges, $G$ serves as a mapping function from the source domain (NYC) to the target domain (LA), implicitly removing the effect of $F$.  

However, merely applying this mechanism to our setting allows for too much freedom in the space of generators $G$.  It could be that the generated dataset $G(X)$ loses all structure associated with $X$, and thus no longer has a use in understanding the data in the source domain!  

To balance this shortcoming, in addition to adopting the domain-adversarial approach, we also add a structure preservation component to measure the similarity of the generated embedding $G(X)$ and the original embedding $X$ by adding cosine distance to the generator loss.  Ideally we would want to preserve the pairwise cosine distance: so for all $x_1, x_2 \in X$ that $\cos(G(x_1), G(x_2)) \approx \cos(x_1, x_2)$.  However, this would induce intractable $p^2$ terms, and the $p$ terms conserving $\cos(G(x),x)$ are a suitable proxy (by triangle inequality).  

\begin{figure}[t]
    \centering
  \includegraphics[width=0.75\linewidth]{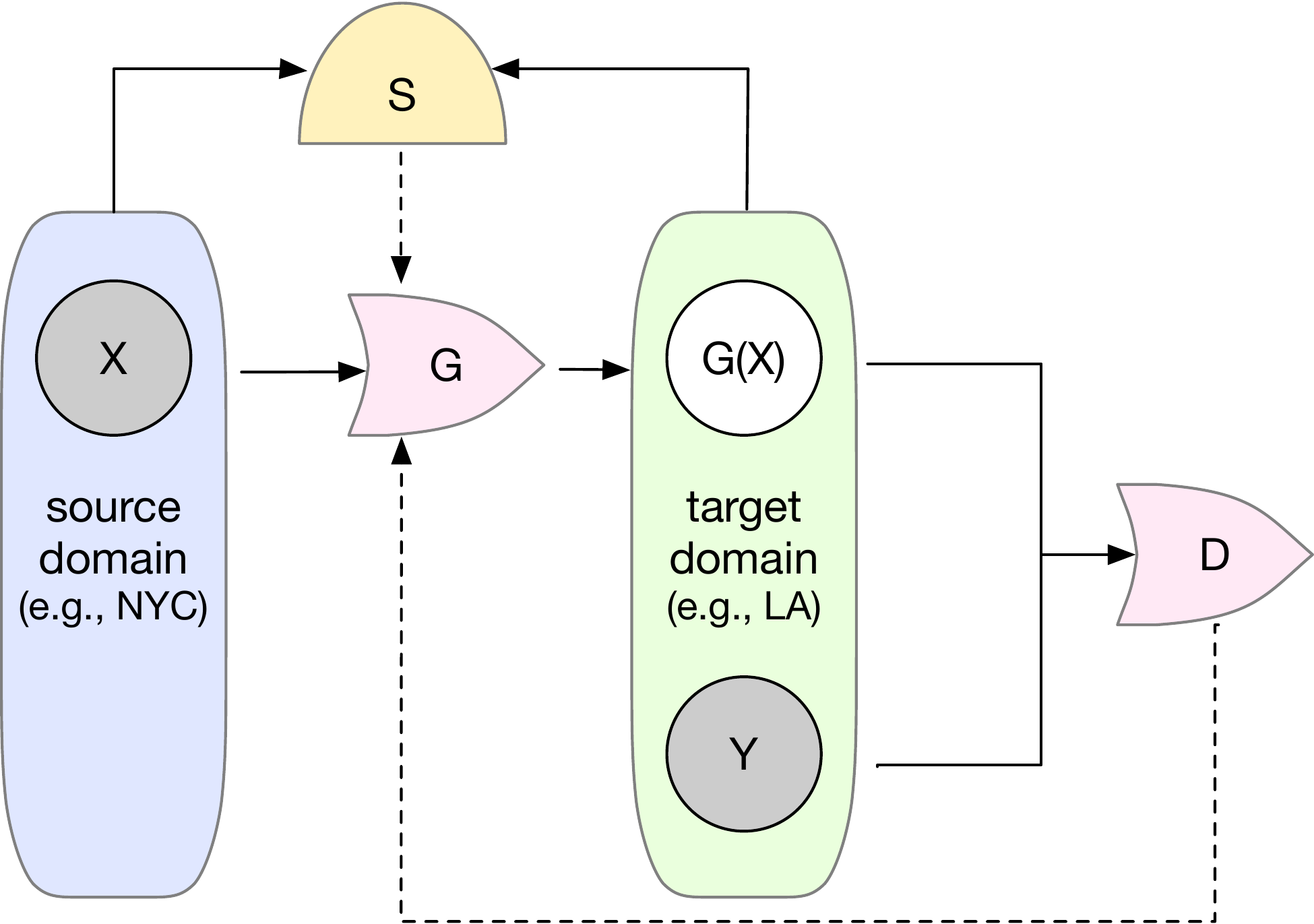}
  \caption{DAN structure of UCAN. $X$ represents an embedding in the source domain, $G(X)$ is the embedding mapped from the source domain to the target domain, and $Y$ represents an embedding in the target domain.
$G$ is the generator, $D$ is the discriminator, and $S$ is the structure preservation component.}
  \label{fig:gan-fig}
%\vspace{-3mm}
\end{figure}

Our proposed method UCAN (Unsupervised Constrained Alignment that is Non-Affine) uses the DAN structure shown in Figure \ref{fig:gan-fig}, where the $D : \mathbb{R}^d \to [0,1]$ and $G : \mathbb{R}^d \to \mathbb{R}^d$ are multilayer perceptrons.  
Ultimately, this is formulated so $D$ and $G$ play the two-player minmax game with value function $V(G,D)$:
%\begin{align}
%    \min_{G}\max_{D}V(D,G) = &\mathbb{E}_{y\sim p_Y(y)}[D(y)] \;\; + \label{eq:1}
%    \\ & \mathbb{E}_{x\sim p_X(x)}[(1-D(G(x))) \label{eq:2}
%    \\ & \phantom{\mathbb{E}_{x\sum p_X(x)}[}+ \mathbf{d}_{cos}(x,G(x))] \label{eq:3}
%\end{align}

\begin{align}
\min_{G}\max_{D}V(D,G) = &\mathbb{E}_{y\sim p_Y(y)}[\log(D(y))]  \label{eq:1}
\\ & +
\mathbb{E}_{x\sim p_X(x)}[\log(1-D(G(x)))]   \label{eq:2}
\\ & +
\alpha \cdot \mathbb{E}_{x\sim p_X(x)}[1-\cos(x,G(x))]   \label{eq:3}
\end{align}

% $X$ represents the original embedding in the source domain, $G(X)$ is the mapped source embedding in the target domain, and $Y$ represents original embedding in the target domain.
%$G$ is the generator, $D$ is the discriminator, and $S$ is the structure preservation component.
%The network of generator and discriminator are both multilayer perceptrons.

%We let $P_D(\text{target}=1 \mid z)$ denote the probability that the discriminator $D$ predicts the vector $z$ is from target domain, and similarly $P_D(\text{target}=0 \mid z)$ is the probability it is predicted to be not from the target domain.  
Since the discriminator tries to separate the target embedding (e.g., LA merchants) and mapped source embedding (e.g., NYC merchants), the discriminator loss function $L_D$ derived from Eq.~(\ref{eq:1}) and Eq.~(\ref{eq:2}), and can be written as: 
%\begin{multline}
\[
L_D =  - \frac{1}{q}\sum_{i=1}^q \log ( D(y_i))) %\\\vspace{-10mm} 
- \frac{1}{p}\sum_{i=1}^p \log ( 1 - D(G(x_i))).
\]
%\end{multline}  
%\begin{multline}
% \begin{split}   L_D = -\frac{1}{q}\sum_{i=1}^q \log P_D(\text{target}=1 \mid y_i) \\ - \frac{1}{p}\sum_{i=1}^p \log  P_D (\text{target}=0 \mid G(x_i))
%\end{multline}

The generator $G$ has two objectives: one is to fool the discriminator (e.g., so $D$ cannot distinguish NYC merchants from LA ones), which makes the discriminator believe that the mapped embeddings are from the target distribution; the other is the structure preservation (S), which is to make the mapped embedding and the original embedding as similar as possible (e.g., so for an NYC merchant $x$, its mapped representation $G(x)$ still retains properties of $x$). 
We use cosine similarity as the similarity measure, as it is the standard similarity optimized in the creation of embeddings. 
We use a loss function $L_G$ with two terms coming separately from Eq.~(\ref{eq:3}) and Eq.~(\ref{eq:2}) which is a proxy for $G(X)$ and $Y$ having a good alignment. It is written as:
%\begin{multline}
\[
L_G = - \alpha\cdot \frac{1}{p}\sum_{i=1}^p \cos(G(x_i), x_i) %\\ \vspace{-10mm} 
- \frac{1}{p}\sum_{i=1}^p \log ( D(G(x_i))).
\]
\section{Experiments}
\label{sec:experiment}
We run an extensive set of experiments to validate our methods. 
To visually demonstrate the idea of our algorithm, we first show the experiment on synthetic datasets.
%and multi-label setting.
Then we further show the effectiveness on three real-world datasets: two public datasets and one industry dataset. We demonstrate two real-world applications for the industry dataset, where removing a location feature by our approach improves the merchant identification accuracy and cross-city restaurant recommendation performance.  
The proposed algorithm significantly outperforms the state-of-the-art unsupervised linear methods on all of the datasets.

We use linear SVM as the classifier to identify the features; it could be replaced by other classifiers. The model is trained on $80\%$ of the data, and the one vs. all AUC is calculated from the remaining $20\%$ of the data. We repeat the experiments 10 times to get the average results. 
For all the experiments, F1 is the feature we want to retain, and F2 is the feature we want to remove.

\subsection{Synthetic Datasets}
Synthetic datasets have two dimensions; each corresponds to one type of binary feature. The x-axis varies with the color feature, and the y-axis varies with the lightness feature. 
The two binary features split the dataset into four parts, and each part is generated by a Gaussian distribution with the same covariance matrix.
In this experiment, we aim to remove the color feature (F2) and retain the lightness feature (F1).

\begin{figure}[t!!]
\centering 
\begin{subfigure}[b]{0.23\textwidth}
    \centering
    \includegraphics[width=\linewidth, page=1]{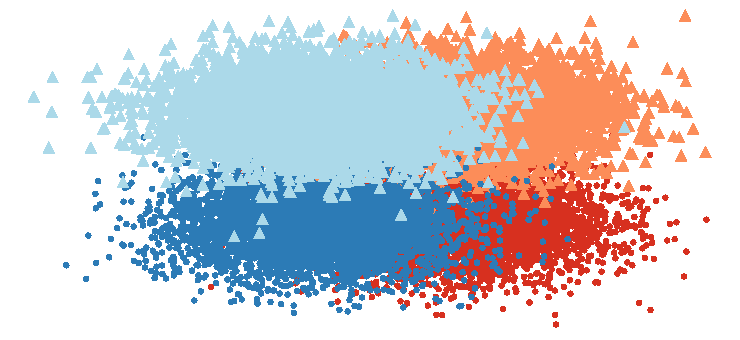}
    %\vspace{-4mm}
    \caption{Raw data}
\end{subfigure}
% \hspace{3mm}
\hfill
\begin{subfigure}[b]{0.23\textwidth}
    \centering
    \includegraphics[width=\linewidth, page=2]{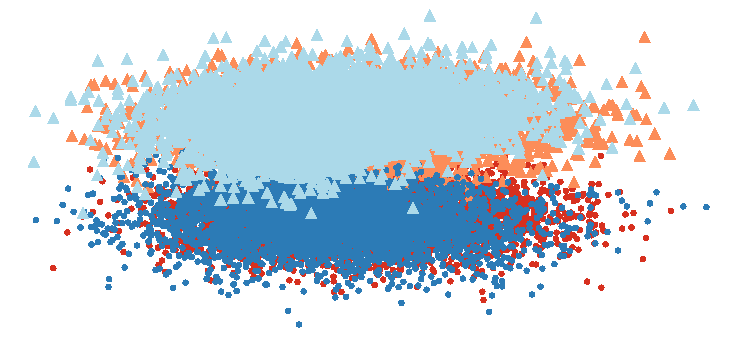}
    %\vspace{-4mm}
    \caption{$\alpha=5$}
\end{subfigure}

\begin{subfigure}[b]{0.23\textwidth}
    \centering
    \includegraphics[width=\linewidth, page=3]{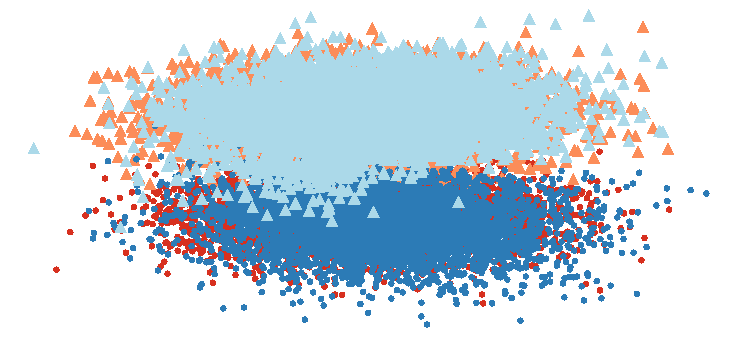}
    %\vspace{-4mm}
    \caption{$\alpha=1$}
\end{subfigure}
% \hspace{3mm}
\hfill
\begin{subfigure}[b]{0.23\textwidth}
    \centering
    \includegraphics[width=\linewidth, page=4]{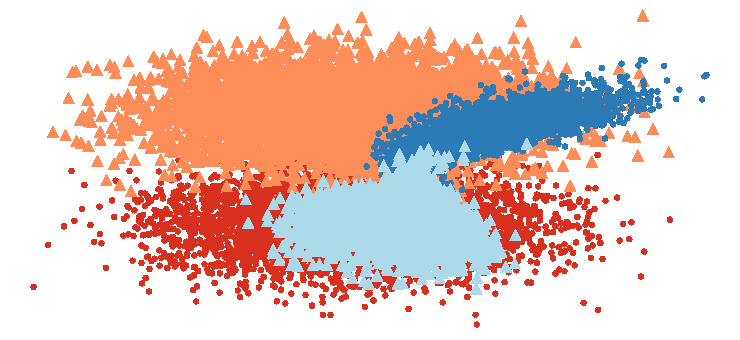}
    %\vspace{-4mm}
    \caption{$\alpha=0$}
\end{subfigure}
% \begin{subfigure}[b]{0.4\textwidth}
%     \centering
%     \includegraphics[width=0.5\linewidth, page=1]{figs/fig2.pdf}
% \end{subfigure}
% \begin{subfigure}[b]{0.4\textwidth}
%     \centering
%     \includegraphics[width=0.5\linewidth, page=2]{figs/fig2.pdf}
% \end{subfigure}
% \includegraphics[width=0.2\textwidth]{figs/synb1_org.png} \hfill
% \includegraphics[width=0.2\textwidth]{figs/synb1_alpha5.png}
% \\ \phantom{12345} (a) original \hfill (b) $\alpha = 5$ \phantom{12345}
% \\
% \includegraphics[width=0.2\textwidth]{figs/synb1_alpha01.png} \hfill
% \includegraphics[width=0.2\textwidth]{figs/synb1_alpha0.png}
% \\  \phantom{12345} (c) $\alpha = 1$ \hfill (b) $\alpha=0$  \phantom{12345} 
    \caption{Balanced synthetic data}
\label{fig:syn_binary_balance}
\end{figure}

\subsubsection{Balanced Dataset}
Based on the color feature, we separate this dataset into two parts, each of which has $10,000$ points.
The original visualization of the dataset is shown in Figure \ref{fig:syn_binary_balance}(a).
To remove the color feature, we map blue points to red points. 
By varying $\alpha$, which controls how much we preserve the structure of the original dataset, the results differ visually and analytically. 
From Table \ref{table_binary_balance}, smaller $\alpha$ ($\alpha > 0$) emphasizes the generator matching $G(X)$ to $Y$, making it more challenging for the discriminator to distinguish the target data and generated data, and thus the smaller AUC on color feature. Lightness feature is resilient to $\alpha$ in $0.1$ to $5$, and retains high AUC (from $0.98$ to $0.97$; see Table \ref{table_binary_balance}).
We can also observe from Figure \ref{fig:syn_binary_balance}(b)(c) that smaller $\alpha$ ($\alpha > 0$) can better align the source to target domain.
However, when $\alpha$ is too small (see $\alpha =0$, so no constraint term), we give too much freedom to the generator without preserving the original data, and thus F2 is not well preserved in Figure \ref{fig:syn_binary_balance}(d), and the AUC of F2 is reduced dramatically. 
The optimal $\alpha$ varies by datasets, but the algorithms are generally robust to this choice.  Empirically, $\alpha = 1$ works well for most datasets (which balances the matching of $G(X)$ and $Y$ with $G(X)$ and $X$), and unless specified, we use $\alpha = 1$ for the remaining experiments.

\begin{table}[t]
%\resizebox{0.95\textwidth}{!}{ % If your table exceeds the column or page width, use this command to reduce it slightly
\caption{Results on Synthetic Dataset. We report the AUC for lightness feature (F1) and color feature (F2).}
\begin{tabular}{l||c|c|c|c}
\hline
 & \multicolumn{2}{c|}{Balanced Data} & \multicolumn{2}{c}{Imbalanced Data} 
\\ & F1\_AUC  & F2\_AUC  & F1\_AUC  & F2\_AUC \\ 
\hline
Raw data  &  0.98& 0.82  &  0.98&  0.82 \\
$\alpha=5$  &  0.97& 0.70  &  0.97&  0.50 \\
$\alpha=1$  &  0.97& 0.60  &  0.97&  0.50 \\
$\alpha=0.5$&  0.97& 0.57  &  0.97&  0.50 \\
$\alpha=0.3$&  0.97& 0.55  &  0.97&  0.50 \\
$\alpha=0.1$&  0.97& 0.52  &  0.97&  0.50 \\
$\alpha=0$  &  0.59& 0.78  &  0.88&  0.50 \\
\hline
\end{tabular}
\label{table_binary_balance}
\end{table}

\subsubsection{Imbalanced Dataset}
\begin{figure}[]
\centering 
%\vspace{-4mm}
\begin{subfigure}[b]{0.23\textwidth}
    \centering
    \includegraphics[width=\linewidth, page=1]{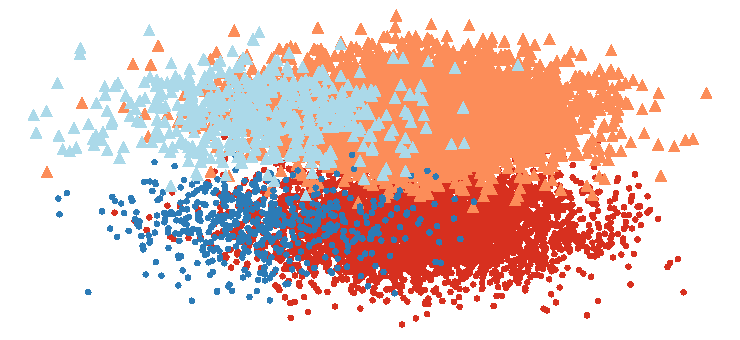}
    %\vspace{-4mm}
    \caption{Raw data}
\end{subfigure}
% \hspace{3mm}
\hfill
\begin{subfigure}[b]{0.23\textwidth}
    \centering
    \includegraphics[width=\linewidth, page=2]{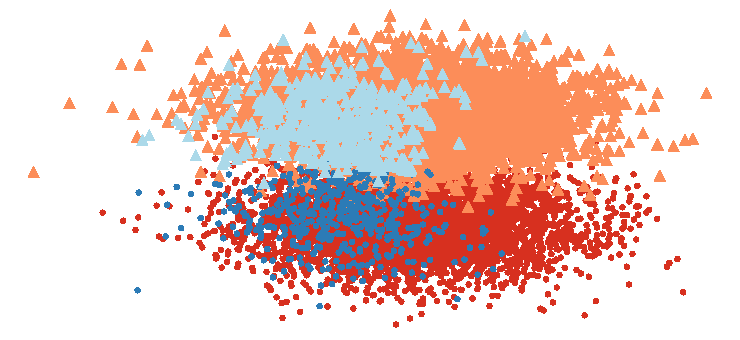}
    %\vspace{-4mm}
    \caption{$\alpha=5$}
\end{subfigure}

\begin{subfigure}[b]{0.23\textwidth}
    \centering
    \includegraphics[width=\linewidth, page=3]{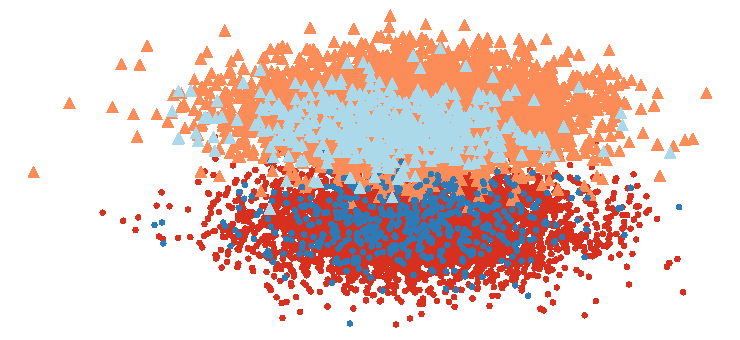}
    %\vspace{-4mm}
    \caption{$\alpha=1$}
\end{subfigure}
% \hspace{3mm}
\hfill
\begin{subfigure}[b]{0.23\textwidth}
    \centering
    \includegraphics[width=\linewidth, page=4]{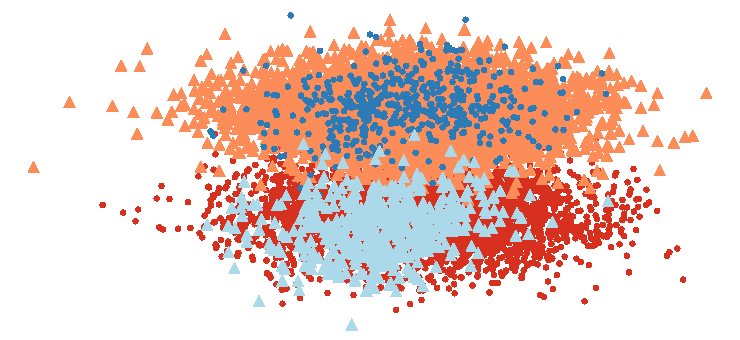}
    %\vspace{-4mm}
    \caption{$\alpha=0$}
\end{subfigure}
% \begin{subfigure}[b]{0.4\textwidth}
%     \centering
%     \includegraphics[width=0.5\linewidth, page=1]{figs/fig2.pdf}
% \end{subfigure}
% \begin{subfigure}[b]{0.4\textwidth}
%     \centering
%     \includegraphics[width=0.5\linewidth, page=2]{figs/fig2.pdf}
% \end{subfigure}
% \includegraphics[width=0.2\textwidth]{figs/synb1_org.png} \hfill
% \includegraphics[width=0.2\textwidth]{figs/synb1_alpha5.png}
% \\ \phantom{12345} (a) original \hfill (b) $\alpha = 5$ \phantom{12345}
% \\
% \includegraphics[width=0.2\textwidth]{figs/synb1_alpha01.png} \hfill
% \includegraphics[width=0.2\textwidth]{figs/synb1_alpha0.png}
% \\  \phantom{12345} (c) $\alpha = 1$ \hfill (b) $\alpha=0$  \phantom{12345} 
    \caption{Imbalanced synthetic data}
\label{fig:syn_binary_imbalance}
\end{figure}

The second synthetic dataset (see Figure \ref{fig:syn_binary_imbalance}) is generated through a similar process except for the number of each set of blue points reduces from $10,000$ to $1,000$; the color feature separates the whole dataset into two imbalanced parts. 
We still attempt to align the blue points to the red points.  Color feature initially has an AUC of $0.88$ and can be reduced to AUC $= 0.5$ with $\alpha=0.1$ through $\alpha = 5$.  When $\alpha=0$, the Lightness feature's AUC remains $0.88$. %For this imbalanced dataset, the feature can be easily reduced by using AUC as the metric. 
%From Figure \ref{fig:syn_binary_imbalance}, visually we observe smaller $\alpha$ ($\alpha > 0$) better maps the source domain points to the target domain.  

\begin{table*}[t]
\caption{Results on airport embedding dataset. O\_ represents the original result, C\_ is the proposed algorithm (UCAN) result, M\_ is the MUSE baseline and U\_ is the UMWE baseline. We report the AUC scores for activity feature (F1) and location feature (F2). }
\centering
\resizebox{\textwidth}{!}{
\begin{tabular}{l||ccc|ccc|ccc|ccc}
\hline
  & O\_F1  & O\_F2 &  O\_F1/O\_F2 & C\_F1 & C\_F2 & C\_F1/C\_F2 & M\_F1  & M\_F2 & M\_F1/M\_F2 & U\_F1  & U\_F2 &  U\_F1/U\_F2\\ 
\hline
Brazil$\to$Europe & 0.80 & 1 & 0.80 & 0.77 & 0.84 & ${\bf0.92}$ & 0.76 & 0.97 & 0.78 & 0.57 & 0.71 & NA \\
Europe$\to$Brazil & 0.80 & 1 & 0.80 & 0.79 & 0.86 & ${\bf0.92}$ & 0.82 & 0.95 & 0.86 & 0.57 & 0.80 & NA \\
Brazil$\to$USA & 0.85 & 1 & 0.85 & 0.85 & 0.61 & ${\bf1.39}$ & 0.83 & 0.98 & 0.85 & 0.5 & 1 & NA \\
USA$\to$Brazil & 0.85 & 1 & 0.85 & 0.88 & 0.91 & ${\bf0.97}$ & 0.87 & 0.93 & 0.94 & 0.52 & 0.51 & NA\\
USA$\to$Europe & 0.84 & 1 & 0.84 & 0.85 & 0.90 & ${\bf0.94}$ & 0.83 & 0.92 & 0.90 & 0.54 & 0.53 & NA \\
Europe$\to$USA & 0.84 & 1 & 0.84 & 0.83 & 0.62 & ${\bf1.34}$ & 0.80 & 0.99 & 0.81 & 0.49 & 0.97 & NA \\
\hline
\end{tabular}
}
\label{table_airport}
\end{table*}

\begin{figure}[]
    \centering
    \begin{subfigure}[b]{0.22\textwidth}
        \centering
        \includegraphics[width=\textwidth]{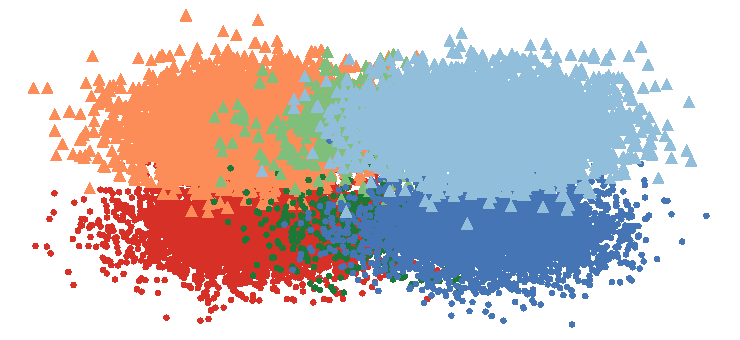}
        \caption{Raw data with }
    \end{subfigure}
    
    \begin{subfigure}[b]{0.22\textwidth}
        \centering
        \includegraphics[width=\textwidth]{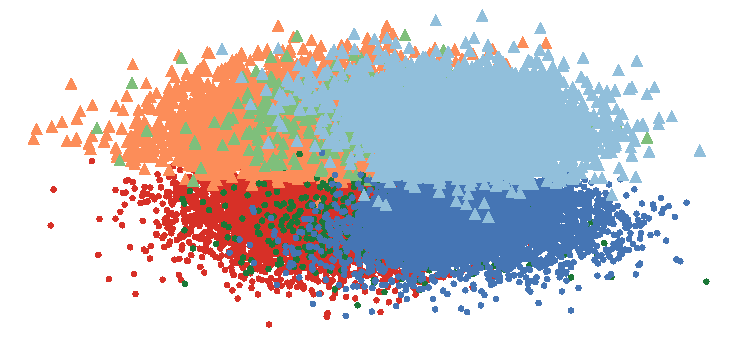}
        \caption{Green as target domain}
    \end{subfigure}
    \begin{subfigure}[b]{0.22\textwidth}
        \centering
        \includegraphics[width=\textwidth]{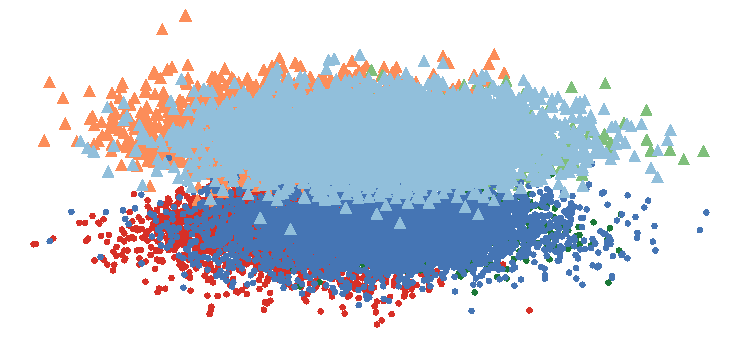}
        \caption{Red as target domain}
    \end{subfigure}
    \caption{Imbalanced multi-label synthetic dataset.}
    \label{fig:syn_multi_imbalance}
\end{figure}

We continue our experiments with the synthetic datasets which have two classes for the lightness feature (light and dark), but now three classes for the color feature (green, red and blue).
The visualization of the multi-label dataset is the same as the binary dataset: the x-axis varies with Feature 1 (F1) which is the color feature, and the y-axis varies with Feature 2 (F2) which is the lightness feature.
To reduce the color feature, we can fix one color as the target domain and the other two colors as the source domains. 
For the balanced dataset, the choice of color as the target domain does not affect the results.
However, for the imbalanced case where there are $1,000$ green points, $10,000$ red points and $10,000$ blue points (illustrated in Figure~\ref{fig:syn_multi_imbalance}(a)), the choice of target domain has a significant impact on the result. When we choose green as the target domain and map red and blue points to green points (Figure~\ref{fig:syn_multi_imbalance}(b)), 
although red and blue points are aligned with green points, they are still not overlapped as desired. This is because blue or red points tend to align along their boundaries due to the structure-preserving components. Although smaller $\alpha$ could solve the issue, the rule of thumb solution is choosing the domain with the most points as the target domain. For example, if we choose red points as the target domain and map green and blue points to the domain of the red points (Figure~\ref{fig:syn_multi_imbalance}(c)), the three-color datasets are well overlapped. 
The color feature are significantly removed with AUC from $0.77$ to $0.64$ (Table~\ref{table_binary_imbalance_multi}).

% \begin{figure}[b]
% \includegraphics[width=0.2\textwidth]{figs/synb3_org.png} \hfill
% \includegraphics[width=0.2\textwidth]{figs/synb3_alpha5.png}
% \\ \phantom{12345} (a) original \hfill (b) $\alpha = 5$ \phantom{12345}
% \\
% \includegraphics[width=0.2\textwidth]{figs/synb3_alpha01.png} \hfill
% \includegraphics[width=0.2\textwidth]{figs/synb3_alpha0.png}
% \\  \phantom{12345} (c) $\alpha = 1$ \hfill (b) $\alpha=0$  \phantom{12345} 
%     \caption{Imbalanced Synthetic data}
% \end{figure}

% \begin{figure}[htb]
%     \centering % <-- added
%     \subfigure[original]{\includegraphics[width=0.15\textwidth]{figs/synm1_org.png}} 
%     \subfigure[Orange as target domain]{\includegraphics[width=0.15\textwidth]{figs/synm1_g1.png}}
%     \subfigure[Blue as target domain]{\includegraphics[width=0.15\textwidth]{figs/synm1_g2.png}}
%     \caption{Imbalanced Synthetic data with binary label}
%     \label{fig:syn_multi_imbalance}
% \end{figure}

\begin{table}[]
\caption{Results on imbalanced synthetic multi-label dataset. We report the AUC for lightness (F1) and color (F2). }
\centering
%\resizebox{0.95\textwidth}{!}{ % If your table exceeds the column or page width, use this command to reduce it slightly
\begin{tabular}{l||c|c}
\hline
 & F1\_AUC  &  F2\_AUC\\ 
\hline
Raw data         &  0.98 &  0.81 \\
Green\_Target   & 0.98  &  0.77 \\
Red\_Target     &  0.98 &  0.64 \\
\hline
\end{tabular}
\label{table_binary_imbalance_multi}
\end{table}

\begin{table*}
\caption{Results on multi-Language embedding for word translation. NN and CSLS are two evaluation approaches. We use precision as metric for $K=1,5,10$. C\_ is the proposed algorithm (UCAN) result, and M\_ is the MUSE baseline method and U\_ is the UMWE baseline method.}
\centering
\resizebox{0.8\textwidth}{!}{ % If your table exceeds the column or page width, use this command to reduce it slightly
\begin{tabular}{l||c|cc|cc|cc|cc|cc}
\hline
 & & en-es & es-en & en-fr & fr-en & en-de & de-en & en-ru & ru-en & en-it & it-en\\ 
\hline
%\multirow{1}{*}{M\_NN\_R} & K@1 & 69.8 & 71.3 & 70.4 & 61.9 & 63.1 &59.6 & 29.1 & 41.5 & NA & NA \\
%\hline
\multirow{3}{*}{M\_NN} & K@1& ${\bf69.07}$ & 64.40 & 53.96 & 61.6 & ${\bf61.02}$ &51.40 & 24.20 & 32.00 & 56.45 & 59.33 \\
                          & K@5 & ${\bf82.93}$ & 78.40 & 65.08 & 75.93& 71.19 & 66.60 & 44.20 & 50.00 & 66.13 & 73.73\\
                          & K@10 & ${\bf86.87}$ & 81.8 & 68.25 & 80.60& 72.88 & 71.93 & 51.40 & 56.20 &  69.35 & 81.13\\
\hline
 \multirow{3}{*}{U\_NN} & K@1&64.13 & 61.73 & 62.47 & 61.27 & 52.40 & 51.73 & ${\bf27.47}$ & 39.13 & 56.73 & 57.27\\
& K@5 & 76.20 & 75.80 & 75.73 & 75.67 & 74.27 & 67.33 & ${\bf52.33}$ & 57.67 & 71.67 & 71.27\\
& K@10 & 79.47 & 79.60 & 79.60 & 79.47 & 79.13 & 71.87 & ${\bf60.87}$ & 63.53 & 76.33 & 75.33\\
\hline
 \multirow{3}{*}{C\_NN} & K@1& 68.93 & ${\bf72.00}$ & ${\bf69.60}$ & ${\bf69.00}$ & 60.27 & ${\bf61.80}$ & 25.73 & ${\bf46.60}$ & ${\bf61.33}$ & ${\bf63.07}$ \\
 & K@5 & 81.26 & ${\bf84.60}$ & ${\bf82.93}$ & ${\bf83.13}$ & ${\bf79.33}$ & ${\bf76.26}$ & 51.33 & ${\bf65.73}$ & ${\bf77.93}$ & ${\bf79.07}$\\
& K@10 & 85.00 & ${\bf88.40}$ & ${\bf85.80}$ & ${\bf87.20}$ & ${\bf83.27}$ & ${\bf80.80}$ & 59.80 & ${\bf70.26}$ &  ${\bf82.87}$ & ${\bf82.53}$\\
\hline
\hline
%\multirow{1}{*}{M\_CSLS\_R} & K@1 & 75.7 & 79.7 & 77.8 & 71.2 & 70.1 &66.4 & 37.2 & 48.1 & 66.2 & 58.7 \\
%\hline
 \multirow{3}{*}{M\_CSLS} & K@1& ${\bf76.00}$ & 71.93 & 68.25 & 69.87& ${\bf71.19}$ & 57.20 & 27.80 & 37.00 & 66.13 & 66.53\\
                             & K@5 & ${\bf86.86}$ & 83.20 & 79.37 & 82.47& 79.66 & 72.87 & 49.93 & 57.67 & 79.03 & 79.20\\
                             & K@10 & ${\bf89.60}$ & 85.87 & 80.95 & 85.40& 81.36 & 77.07 & 56.60 & 63.53 & 82.26 & 82.33\\
\hline
 \multirow{3}{*}{U\_CSLS} & K@1& 70.07 & 68.67 & 69.00 & 68.93 & 59.47 & 58.80 & ${\bf32.33}$ & 45.80 & 63.53 & 65.13\\
 & K@5 & 81.20 & 81.27 & 82.20 & 81.67 & 78.60 & 73.47 & ${\bf58.33}$ & 64.33 & 77.27 & 77.07\\
 & K@10 & 84.60 & 84.20 & 84.87 & 84.93 & 83.20 & 76.93 & ${\bf66.27}$ & 69.93 & 82.00 & 80.33\\
\hline
\multirow{3}{*}{C\_CSLS} & K@1& 72.00 & ${\bf77.60}$ & ${\bf74.93}$ & ${\bf75.73}$& 64.47 & ${\bf65.33}$ & 30.00 & ${\bf50.20}$ & ${\bf66.93}$ & ${\bf69.20}$\\
 & K@5 & 84.46 & ${\bf87.60}$ & ${\bf86.40}$ & ${\bf87.47}$& ${\bf81.27}$ & ${\bf78.60}$ & 57.20 & ${\bf70.20}$ & ${\bf81.80}$ & ${\bf82.73}$\\
 & K@10 & 87.13 & ${\bf90.00}$ & ${\bf88.46}$ & ${\bf90.33}$& ${\bf85.60}$ & ${\bf83.33}$ & 64.26 & ${\bf74.40}$ & ${\bf85.47}$ & ${\bf86.07}$\\
 \hline
\end{tabular}
}
\label{table_nn_lang}
\end{table*}

\subsection{Real Datasets}
We use three real-world datasets to demonstrate the effectiveness of the proposed algorithm. 
Since our goal is to retain the F1 feature and remove the F2 feature,  if AUC score of F1 is not reduced more than $10\%$, the ratio score (AUC score of F1)/(AUC score of F2) is being calculated. A higher score means better performance. If AUC score of F1 is reduced more than $10\%$, the ratio score is set to NA.
O\_ represents the original result, C\_ is the proposed algorithm result, M\_ (MUSE~\cite{Conneau2018}) and U\_ (UMWE~\cite{ChenC18}) are the baseline results. 
For MUSE and UMWE, we use the default setting, where we run $5$ million iterations with batch size $32$, the number of refinements is $5$ for the airport and merchant embedding dataset and $0$ for the multi-language embedding dataset. For our algorithm, we set the number of iterations as $1$ million, batch size as $32$ for multi-language embedding dataset;  the number of iterations as $30k$ and batch size as $512$ for all the other experiments.

\subsubsection{Airport Embedding Dataset}
The air-traffic networks dataset includes three air-traffic networks \cite{figueiredo2017struc2vec}:
Brazil, Europe and the USA: each node is an airport, and each edge shows the existence of commercial flights between airports.  
We consider two features: the level of activity (F1) with $4$ classes, and country location (F2) with $3$ classes.  
%Their class labels are assigned based on the level of
%activity measured by flights or people that passed through the airports. 
We apply the DEMO-net \cite{wu2019net}, which is a degree-specific graph neural network for node and graph classification, to train the model and get embeddings of all the nodes.
Since the three networks are trained separately, if we view all the embeddings as one dataset, the location feature (F2) is definitely embedded.  
With the level of activity as the label used during the whole training process, the level of activity (F1) is also strongly represented in the embeddings.   
We use pairwise classification henceforth. From Table \ref{table_airport}, the original features are labeled as O\_F1 and O\_F2, for all pairs, the AUC for location feature is $1$ and AUC for the activity feature is around $0.8$.
Our approach (labeled as C\_F1 and C\_F2) can reduce the weight of the location feature (F2) by about $9\% - 39\%$ for different pairs with minimal degradation of activity (F1).  
The result of the baseline methods MUSE, UMWE are labeled as M\_F1 and M\_F2 \& U\_F1 and U\_F2. Compared with our algorithm, MUSE and UMWE do not perform well. The F1/F2 score only marginally increases for MUSE and not at all for UMWE, whereas our algorithm's ratio score increases about $15\%$ to $63\%$.  We mark U\_F1/U\_F2 as NA since U\_F1 is degraded so much that this measure is not useful.  
%\jeff{Why is U\_F1/U\_F2 marked as NA?}

\subsubsection{Multi-language Embedding}
%For multi-language embedding, ``language'' is feature we want to remove, and meaning of the words from different languages is the feature we want to retain. 
For multi-language embedding, our goal is to remove the ``language'' feature, and retain meanings of the words from different languages.  
%We adopt a high-quality dictionaries of up to $100k$ pairs of words using an Facebook internal translation tool to do the evaluation~\cite{Conneau2018}.
We use unsupervised word vectors that were trained using fastText\footnote{ https://github.com/facebookresearch/fastText}~\cite{bojanowski2017enriching}. 
These correspond to monolingual embeddings of dimension $300$ trained on Wikipedia corpora.
The languages focused on in our experiments are English (en), Spanish (es), French (fr), German (de), Russian (ru) and Italian (it). English is the source domain when translating English to other languages, while other languages are target domains. When we translate other languages into English, English is the target domain, and other languages are the source domains. 
We use the standard K-nearest neighbor (NN) and Cross-Lingual Similarity Scaling (CSLS)~\cite{Conneau2018} as the evaluation approaches.  %, where CSLS can attenuate the hubness problem in high-dimensional spaces to calculate nearest neighbors. 
We measure how many times one of the correct translations of a source word are retrieved, and report the precisions for $K = 1, 5, 10$ in Table \ref{table_nn_lang}.  
%Although CSLS reports higher number than the NN, the conclusion from the two metrics are the same.
Our algorithm outperforms MUSE and UWME in most language pairs, except for MUSE on English to Spanish, and UWME on English to Russian. Our algorithm achieves the best result in the other $8$ pairs and is nearly best on the $2$ exceptions. 
%Experiments on five more languages can be found in Appendix \ref{sec:multi-lang-more}

We also demonstrate the results on languages which are less similar to English, including Greek (el), Vietnamese (vi), Arabic (ar), Czech (cs) and Dutch (nl).
As shown in Table \ref{table_nn_lang_more}, our model can outperform MUSE and UMWE on all the language pairs. 
For language pair el-en, en-vi, vi-en, ar-en, MUSE cannot find a valid mapping. Thus both NN and CSLS results are close to $0$. 
This is probably because el, vi and ar are from different language families as en. 
UMWE can overcome this issue by mapping them at the same time, but both NN and CSLS are not as good as our model.
In conclusion, our model UCAN is very robust and can extend to any language family, while MUSE is limited to work on languages similar to English.

\begin{table*}[]
\caption{Results on multi-Language embedding for word translation. NN and CSLS are two evaluation approaches. We use precision as metric for $K=1,5,10$. C\_ is the proposed algorithm (UCAN) result, and M\_ is the MUSE baseline method and U\_ is the UMWE baseline method.}
\centering
\resizebox{0.85\textwidth}{!}{ % If your table exceeds the column or page width, use this command to reduce it slightly
\begin{tabular}{l||c|cc|cc|cc|cc|cc}
\hline
 &  & en-el & el-en & en-vi & vi-en & en-ar & ar-en & en-cs & cs-en & en-nl & nl-en\\ 
\hline
%\multirow{1}{*}{M\_NN\_R} & K@1 & 69.8 & 71.3 & 70.4 & 61.9 & 63.1 &59.6 & 29.1 & 41.5 & NA & NA \\
%\hline
\multirow{3}{*}{M\_NN} &K@1 & 13.87 & 0.00 & 0.00 & 0.07 & 12.47 & 0.00 & 24.73 & 41.67 & 49.07 & 45.53 \\
&K@5 & 31.13 & 0.00 & 0.13 & 0.07 & 26.33 & 0.07 & 42.00 & 58.33 & 67.80 & 61.87 \\
&K@10 & 38.33 & 0.00 & 0.33 & 0.07 & 32.27 & 0.07 & 49.67 & 64.40 & 73.33 & 67.00 \\
\hline
 \multirow{3}{*}{U\_NN} &K@1 & 18.73 & 26.27 & 3.27 & 1.87 & 14.60 & 22.49 & 26.87 & 31.13 & 43.07 & 33.67 \\
&K@5 & 36.47 & 44.80 & 8.40 & 5.60 & 32.00 & 38.35 & 46.53 & 48.00 & 60.20 & 48.93 \\
&K@10 & 43.27 & 50.53 & 10.73 & 8.53 & 38.87 & 44.51 & 53.20 & 53.07 & 66.07 & 54.13 \\
\hline
 \multirow{3}{*}{C\_NN} &K@1 & ${\bf 28.20 }$ & ${\bf 41.47 }$ & ${\bf 12.13 }$ & ${\bf 31.93 }$ & ${\bf 19.97 }$ & ${\bf 34.80 }$ & ${\bf 28.33 }$ & ${\bf 49.73 }$ & ${\bf 60.00 }$ & ${\bf 61.67}$ \\
&K@5 & ${\bf 46.60 }$ & ${\bf 59.93 }$ & ${\bf 23.27 }$ & ${\bf 45.33 }$ & ${\bf 41.45 }$ & ${\bf 51.74 }$ & ${\bf 50.27 }$ & ${\bf 67.73 }$ & ${\bf 76.40 }$ & ${\bf 75.60}$ \\
&K@10 & ${\bf 52.53 }$ & ${\bf 65.27 }$ & ${\bf 27.67 }$ & ${\bf 49.60 }$ & ${\bf 48.67 }$ & ${\bf 57.50 }$ & ${\bf 58.93 }$ & ${\bf 71.60 }$ & ${\bf 80.60 }$ & ${\bf 79.27}$ \\
\hline
\hline
%\multirow{1}{*}{M\_CSLS\_R} & K@1 & 75.7 & 79.7 & 77.8 & 71.2 & 70.1 &66.4 & 37.2 & 48.1 & 66.2 & 58.7 \\
%\hline
 \multirow{3}{*}{M\_CSLS} &K@1 & 22.67 & 0.00 & 0.00 & 0.00 & 15.00 & 0.00 & 31.07 & 47.53 & 59.67 & 55.33 \\
&K@5 & 42.93 & 0.00 & 0.07 & 0.07 & 32.60 & 0.07 & 52.80 & 65.53 & 77.27 & 70.93 \\
&K@10 & 50.80 & 0.00 & 0.07 & 0.07 & 39.00 & 0.07 & 60.07 & 70.67 & 82.73 & 75.00 \\
\hline
 \multirow{3}{*}{U\_CSLS} &K@1 & 25.20 & 34.47 & 6.27 & 4.20 & 18.47 & 28.45 & 32.93 & 38.53 & 50.67 & 43.67 \\
&K@5 & 44.00 & 54.33 & 13.07 & 10.93 & 37.20 & 46.92 & 53.67 & 55.67 & 67.60 & 57.93 \\
&K@10 & 50.33 & 59.07 & 16.27 & 14.93 & 44.07 & 52.95 & 61.07 & 60.20 & 72.67 & 63.33 \\
\hline
\multirow{3}{*}{C\_CSLS} &K@1 & ${\bf 32.40 }$ & ${\bf 45.80 }$ & ${\bf 21.80 }$ & ${\bf 38.33 }$ & ${\bf 23.93 }$ & ${\bf 36.81 }$ & ${\bf 34.13 }$ & ${\bf 53.60 }$ & ${\bf 67.47 }$ & ${\bf 68.67}$ \\
&K@5 & ${\bf 51.53 }$ & ${\bf 64.27 }$ & ${\bf 33.93 }$ & ${\bf 50.26 }$ & ${\bf 45.60 }$ & ${\bf 54.75 }$ & ${\bf 57.20 }$ & ${\bf 69.80 }$ & ${\bf 82.07 }$ & ${\bf 80.60}$ \\
&K@10 & ${\bf 57.87 }$ & ${\bf 69.47 }$ & ${\bf 38.33 }$ & ${\bf 54.86 }$ & ${\bf 52.28 }$ & ${\bf 59.71 }$ & ${\bf 64.27 }$ & ${\bf 74.47 }$ & ${\bf 85.53 }$ & ${\bf 83.20}$ \\
 \hline
\end{tabular}
}
\label{table_nn_lang_more}
\end{table*}

\subsubsection{Merchant Embedding Dataset}
% In this section, we use a real-world application as a downstream task to evaluate the effectiveness of the embedding mapping algorithm for the merchant embedding dataset. 
% The merchant embedding dataset is generated from a real-world transaction dataset. 
% The embedding is generated by Word2vec~\cite{MSC13,du2019}, where each merchant is treated as a word and each customer as a document. 
% The embedding is trained on all available US transaction data. 
% We detect three prominent features in this dataset: location, number of transactions (frequency), and merchant category code (MCC), where the location is the feature to be removed, and the other features are to be retained. 
% Similar to the airport dataset, we use pair-wise mapping to demonstrate the results. 
% In Table \ref{table_merchant}, F1 is the MCC feature, and F2 is the location feature. 
% Both our algorithm and MUSE can significantly reduce the location feature while retaining the MCC feature; our algorithm significantly outperforms MUSE across all the location pairs.  
% By observing the F1/F2 ratio, our method clearly outperforms MUSE; the details are deferred to the Supplement.  

This embedding dataset is generated from a real-world transaction dataset from a well-known global payment company involving $70$ million merchants and $260$ million customers from December 1, 2017 to June 30, 2019. 
The merchant embedding is generated by Word2vec~\cite{MSC13,du2019},  where each merchant is treated as a word and each customer as a document. 
The embedding is trained on all available US transaction data. 
For this experiment, we focus on the merchant embedding dataset in four areas: Los Angeles (LA), San Francisco (SF), Chicago (CHI), and Manhattan (MAN). 
% We choose the most frequent merchant category code (MCC) as $50$ to filter the embeddings. 
% After filtering, the number of merchants is $136{,}125$, $123{,}896$, $107{,}322$, and $187{,}407$ respectively. 
We detect three prominent features in this dataset: location, frequency, and merchant category code (MCC), where the location is the feature to be removed, and the other features are retained. 
Similar to the airport dataset, we use pair-wise mapping to demonstrate the results. 
In Table~\ref{table_merchant}, F1 is the MCC feature, and F2 is the location feature. 
Both our algorithm and baseline methods can reduce the location feature while retaining the MCC feature; our algorithm significantly outperforms MUSE and UMWE across all the location pairs.  
By observing the F1/F2 ratio score, our method clearly outperforms MUSE and UMWE.

Apart from MCC feature, we can also retain the frequency feature (F3) when removing the location feature. We show the results in Table \ref{table_merchant_frequency}. Both UCAN (our algorithm) and MUSE can significantly reduce the location feature while retaining the frequency feature; UCAN significantly outperforms MUSE across all the location pairs. F3/F2 is increased from $10\%$ to $50\%$ under the MUSE algorithm, while UCAN increases from $43\%$ to $78\%$.
%the details regarding frequency are deferred to Appendix~\ref{sec:mrch_freq}. 

\begin{table*}[ht]
\caption{Results on Merchant Embedding Dataset. O\_ represents the original result, C\_ is the proposed algorithm (UCAN) result, M\_ is the MUSE baseline and  U\_ is the UMWE baseline method. We report the AUC scores for MCC feature (F1) and location feature (F2).}
\centering
\resizebox{0.9\textwidth}{!}{ % If your table exceeds the column or page width, use this command to reduce it slightly
\begin{tabular}{l||ccc|ccc|ccc|ccc}
\hline
  & O\_F1  & O\_F2 &  O\_F1/O\_F2 & C\_F1 & C\_F2 & C\_F1/C\_F2 & M\_F1  & M\_F2 & M\_F1/M\_F2 & U\_F1  & U\_F2 & U\_F1/U\_F2\\ 
\hline
LA$\to$SF   & 0.64 & 0.88 & 0.73 & 0.62 & 0.59 & ${\bf1.05}$ & 0.62 & 0.63 & 0.98 & 0.62 & 0.77 & 0.81\\
SF$\to$LA   & 0.64 & 0.88 & 0.73 & 0.62 & 0.61 & ${\bf1.02}$ & 0.61 & 0.78 & 0.78 & 0.62 & 0.75 & 0.83\\
LA$\to$CHI  & 0.58 & 0.95 & 0.61 & 0.57 & 0.57 & ${\bf1.00}$ & 0.58 & 0.71 & 0.82 & 0.57 & 0.79 & 0.72\\
CHI$\to$LA  & 0.58 & 0.95 & 0.61 & 0.56 & 0.57 & ${\bf0.98}$ & 0.57 & 0.63 & 0.90 & 0.57 & 0.75 & 0.76\\
SF$\to$CHI  & 0.65 & 0.93 & 0.70 & 0.63 & 0.59 & ${\bf1.07}$ & 0.61 & 0.79 & 0.77 & 0.64 & 0.82 & 0.78\\
CHI$\to$SF  & 0.65 & 0.93 & 0.70 & 0.63 & 0.56 & ${\bf1.13}$ & 0.63 & 0.66 & 0.95 & 0.64 & 0.76 & 0.84\\
MAN$\to$LA  & 0.61 & 0.90 & 0.68 & 0.59 & 0.54 & ${\bf1.09}$ & 0.59 & 0.82 & 0.72 & 0.60 & 0.75 & 0.80\\
LA$\to$MAN  & 0.61 & 0.90 & 0.68 & 0.59 & 0.54 & ${\bf1.09}$ & 0.59 & 0.60 & 0.98 & 0.59 & 0.73 & 0.81\\
MAN$\to$SF  & 0.66 & 0.88 & 0.75 & 0.64 & 0.53 & ${\bf1.21}$ & 0.64 & 0.62 & 1.03 & 0.64 & 0.71 & 0.90\\
SF$\to$MAN  & 0.66 & 0.88 & 0.75 & 0.64 & 0.53 & ${\bf1.21}$ & 0.64 & 0.62 & 1.03 & 0.65 & 0.68 & 0.96\\
MAN$\to$CHI & 0.62 & 0.93 & 0.67 & 0.60 & 0.52 & ${\bf1.15}$ & 0.59 & 0.80 & 0.74 & 0.61 & 0.80 & 0.76\\
CHI$\to$MAN & 0.62 & 0.93 & 0.67 & 0.60 & 0.52 & ${\bf1.15}$ & 0.61 & 0.61 & 1.00 & 0.62 & 0.75 & 0.83\\
\hline
\end{tabular}
}
\label{table_merchant}
\end{table*}

\begin{table*}[]
\caption{Results on Merchant Embedding Dataset. O\_ represents the original result, C\_ is the proposed algorithm (UCAN) result, and M\_ is the MUSE baseline. We report the AUC scores for frequency feature (F3) and location feature (F2).}
\centering
\resizebox{0.9\textwidth}{!}{ % If your table exceeds the column or page width, use this command to reduce it slightly
\begin{tabular}{l||ccc|ccc|ccc|ccc}
\hline
  & O\_F3  & O\_F2 &  O\_F3/O\_F2 & C\_F3 & C\_F2 & C\_F3/C\_F2 & M\_F3  & M\_F2 & M\_F3/M\_F2 & U\_F3  & U\_F2 & U\_F3/U\_F2\\ 
\hline
LA$\to$SF    & 0.57 & 0.88 & 0.65 & 0.58 & 0.59 & ${\bf0.98}$ & 0.57 & 0.63 & 0.90 & 0.6 & 0.77 & 0.89 \\
SF$\to$LA    & 0.57 & 0.88 & 0.65 & 0.57 & 0.61 & ${\bf0.93}$ & 0.57 & 0.78 & 0.73 & 0.61 & 0.75 & 0.81\\
LA$\to$CHI   & 0.59 & 0.95 & 0.62 & 0.59 & 0.57 & ${\bf1.04}$ & 0.59 & 0.71 & 0.83 & 0.59 & 0.79 & 0.75\\
CHI$\to$LA   & 0.57 & 0.95 & 0.60 & 0.57 & 0.57 & ${\bf1.00}$ & 0.56 & 0.63 & 0.89 & 0.57 & 0.75 & 0.76 \\
SF$\to$CHI   & 0.57 & 0.93 & 0.61 & 0.57 & 0.59 & ${\bf0.97}$ & 0.57 & 0.79 & 0.72 & 0.62 & 0.82 & 0.76\\
CHI$\to$SF   & 0.59 & 0.93 & 0.63 & 0.59 & 0.56 & ${\bf1.05}$ & 0.59 & 0.66 & 0.89 & 0.59 & 0.76 & 0.78\\
MAN$\to$LA  & 0.57 & 0.9 & 0.63 & 0.56 & 0.54 & ${\bf1.04}$ & 0.57 & 0.82 & 0.70 & 0.59 & 0.75 & 0.79\\
LA$\to$MAN  & 0.57 & 0.9 & 0.63 & 0.57 & 0.54 & ${\bf1.06}$ & 0.56 & 0.6 & 0.93 & 0.58 & 0.73 & 0.79\\
MAN$\to$SF  & 0.58 & 0.88 & 0.66 & 0.59 & 0.53 & ${\bf1.11}$ & 0.58 & 0.62 & 0.94 & 0.6 & 0.71 & 0.85\\
SF$\to$MAN  & 0.59 & 0.88 & 0.67 & 0.58 & 0.53 & ${\bf1.09}$ & 0.59 & 0.62 & 0.95 & 0.6 & 0.68 & 0.88\\
MAN$\to$CHI  & 0.59 & 0.93 & 0.63 & 0.58 & 0.52 & ${\bf1.12}$ & 0.59 & 0.8 & 0.74 & 0.59 & 0.80 & 0.74\\
CHI$\to$MAN  & 0.59 & 0.93 & 0.63 & 0.58 & 0.52 & ${\bf1.12}$ & 0.58 & 0.61 & 0.95 & 0.56 & 0.75 & 0.75\\
\hline
\end{tabular}
}
\label{table_merchant_frequency}
\end{table*}

\subsection{Downstream Task: On Merchant Data Set}
\subsubsection{False Merchant Identity Detection}
In this section, we use a real-world application as a downstream task to evaluate the effectiveness of the embedding mapping algorithm for the merchant embedding dataset. 
Credit/debit card payment volume has proliferated in recent years with the rapid growth of small businesses and online shops. 
When processing these payment transactions, recognizing each merchant's real identity (i.e., merchant category) is vital to ensure the integrity of payment processing systems.
For example, a high-risk merchant may pretend to be in a low-risk merchant category by reporting a fake merchant category to avoid higher processing fees associated with risky categories.
Specific business type (i.e., gambling) is only permitted in some regions and territories.
A merchant could report a false merchant category to avoid scrutiny from banks and regulators.

\begin{figure}
    \centering
  \includegraphics[width=0.8\linewidth]{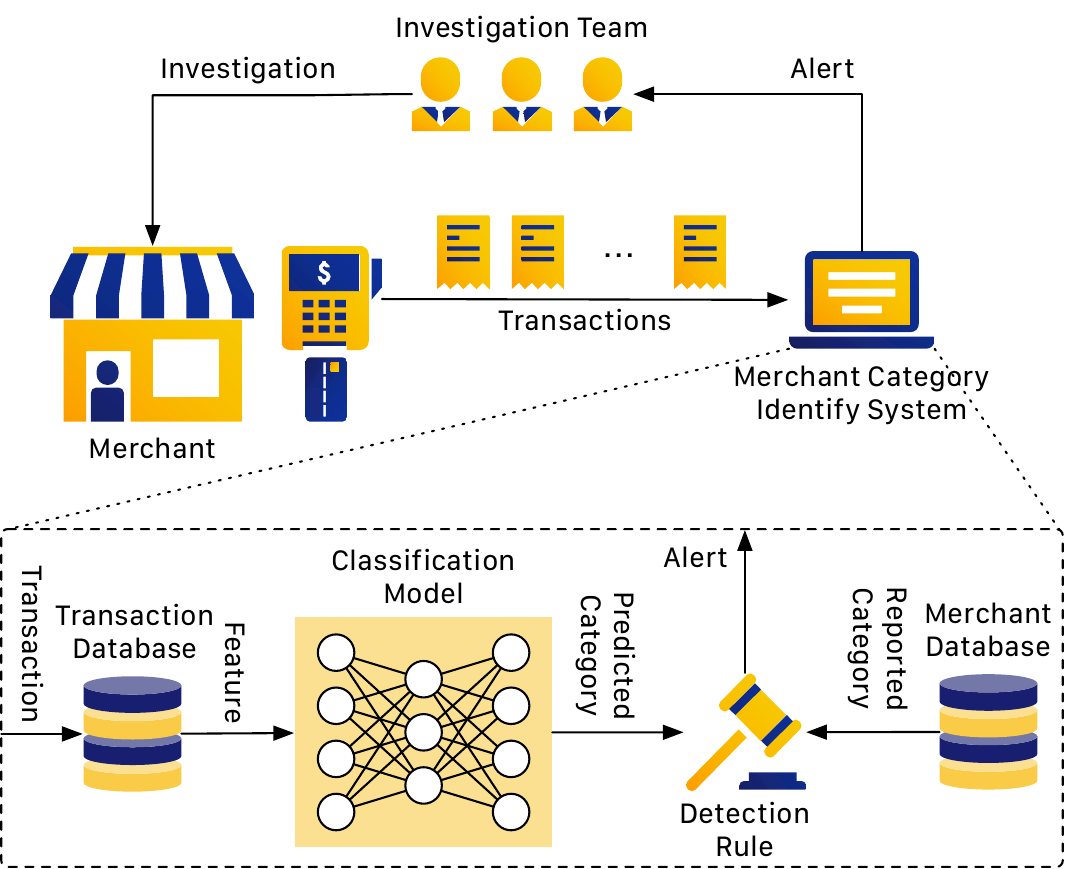}
  \includegraphics[width=0.8\linewidth]{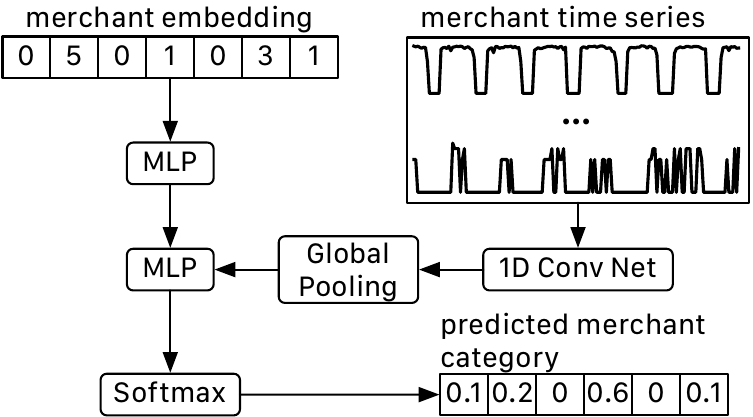}
  \caption{Overall system design and architecture design for our merchant category classifier.
  MLP stand for multilayer perceptron and the 1D convolutional net uses a design similar to the temporal convolutional network.}
  \label{fig:classifier}
  %\vspace{-3mm}
\end{figure}

Accurate embeddings are an essential part of our merchant false identity detection~\cite{yeh2020merchant}.  
The merchant category identification system monitors the transactions of each merchant and notifies the investigation team whenever the identified merchant mismatches with the merchant's self-reported category.
As shown in the architecture design of our classification model (Figure~\ref{fig:classifier}), the two types of features we used for the classification model are each merchant's embedding (i.e., learned through Word2vec based on~\cite{du2019,yeh2020towards}) and the time series capturing each merchant's transaction behavior.
However, the major information captured by the Word2vec-based merchant embedding is the geological location of the merchant.
Thus, attenuating the dominant geolocation component should enhance the classification performance from the embeddings.

\begin{table}[]
\caption{Performance of the merchant category classification model.}
\centering
\resizebox{\linewidth}{!}{ 
\begin{tabular}{l||ccccc}
\hline
         & Micro F1 & Macro F1 & Hit@3  & Hit@5  \\ \hline
Raw data & 0.2884   & 0.2352   & 0.4453 & 0.5154 \\
MUSE     & 0.3142   & 0.1836   & 0.4952 & 0.5931 \\
\textbf{UCAN} (ours)     & \textbf{0.3338}   & \textbf{0.2742}   &  \textbf{0.5592} & \textbf{0.6543} \\ \hline
\end{tabular}
}
\label{table_mrch_class}
\end{table}

To test the performance of different embedding mapping algorithms, we train our model with Los Angeles's merchants and test the model on San Francisco's merchants.
This way, we can directly examine the difference between merchant embeddings contaminated with geological location information and merchant embeddings free of geological location information.
The number of merchant categories in our dataset is 50.  
Table~\ref{table_mrch_class} shows result of classification experiment.
We use conventional classification performance measurements like micro f1, macro f1, hit rate at 3, and hit rate at 5.
As expected, attenuating the geological location information is crucial for learning a more accurate classifier.
The raw embedding performs the worst in almost all performance measurements compared to both MUSE and UCAN.
MUSE general performs better than the raw embedding, but the macro f1 is much worse compared to the raw embedding.
Lastly, our proposed UCAN embedding mapping algorithm overall has the best performance across all performance measurements compared to the baseline method.

\begin{figure}
\centering
  \includegraphics[width=\linewidth]{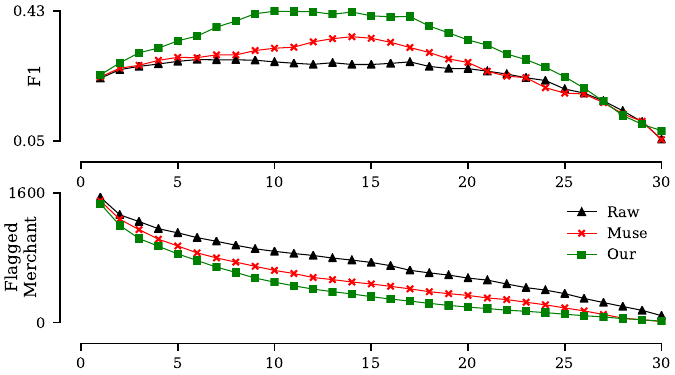}
  \caption{The performance of the merchant category identification system.}
  \label{fig:real_exp}
\end{figure}

Since our goal is to detect merchants with false merchant categories instead of classification, we further evaluate the accuracy of the complete detection system.
In our system, the particular detection rule we used is: if a merchant's self-reported merchant category is not within the top-$k$th most likely merchant category, the merchant will be reported as a suspicious merchant where $k$ is an adjustable threshold for our detection system.
To perform the evaluation, we randomly select $10\%$ of the test merchant and randomly change their self-reported merchant category to emulate the process of one merchant faking its merchant identity.
As we vary the detection threshold, we report the f1 score and the number of suspicious flagged merchants.
The experiment result is shown in Figure~\ref{fig:real_exp}.
Our attenuating method can generate better quality embeddings comparing to other methods.
The resulting merchant category identification system can capture suspicious merchants more accurately and also report less suspicious merchants to the investigation team than the system using raw or MUSE processed embeddings.

%See additional results in
%Additional experiment result is presented in 
%Appendix ~\ref{sec:mcc-result}.

\subsubsection{Cross-City Restaurant Recommendations}
\begin{figure}
    \centering
  \includegraphics[width=0.8\linewidth]{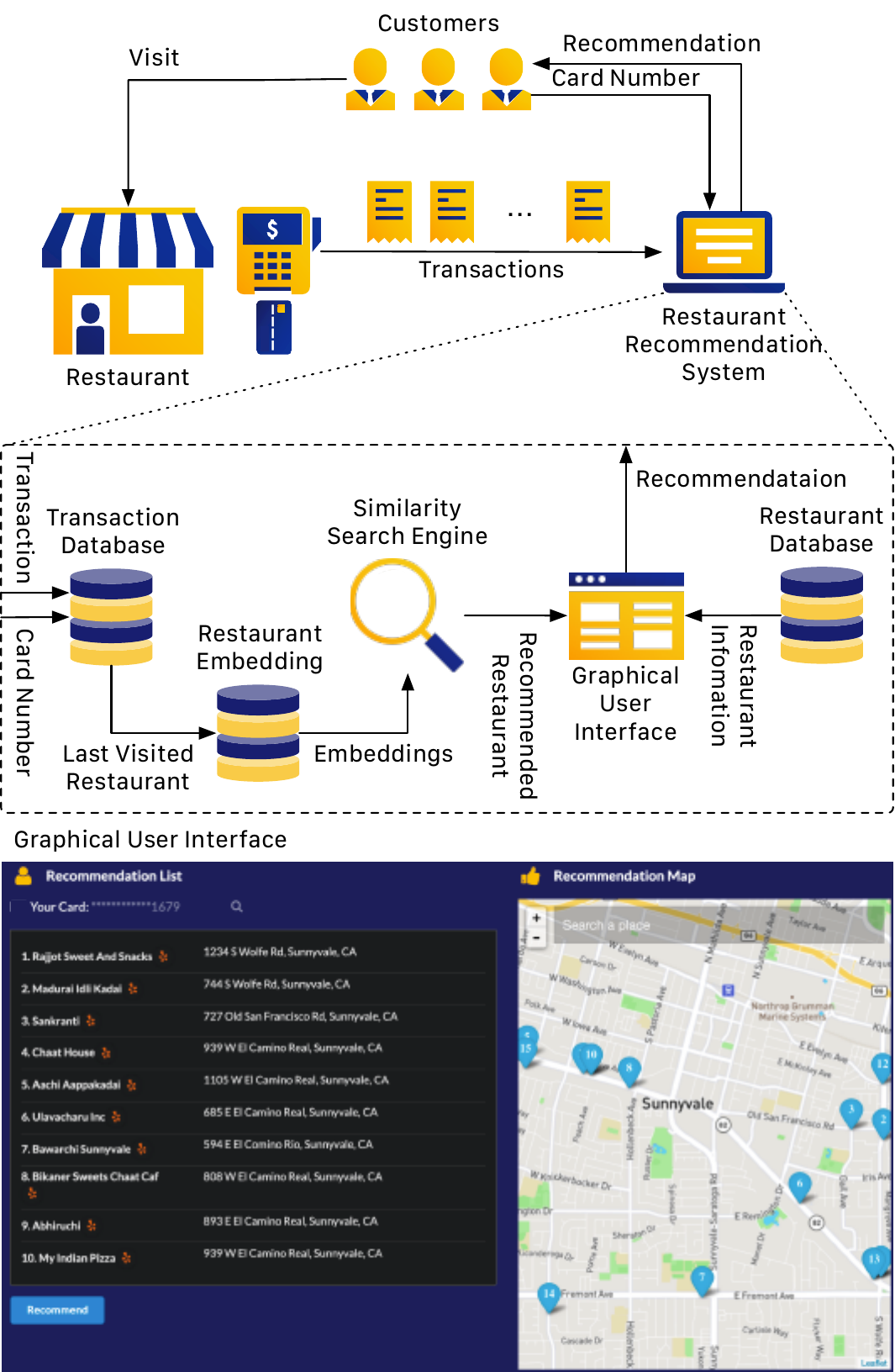}
  \caption{Overall system design for our collaborative filtering-based restaurant recommendation system.
  An example of the graphical user interface in action is shown below.}
  \label{fig:restaurant}
  %\vspace{-3mm}
\end{figure}

Another application under merchant embeddings is to recommend restaurants to a customer across different cities based on the customer's historical transactions in the home city.
We evaluate the effectiveness of removing the location feature using a real-world restaurant recommendation system~\cite{bendre2011gpr}. For this evaluation, we choose the recommendation system's collaborative filtering model, as its performance is heavily dependent on the quality of the embedding. 
Figure~\ref{fig:restaurant} shows how different components of the recommendation system interact with each other during inference time.
For a specified area (zip code), the recommendation model ranks the restaurants on how similar they are to the restaurants visited by a customer previously and uses the similarity score to rank the restaurants. 
For this experiment, we sample around $10$K customers having restaurant transactions from May 2019 to Oct 2019 in SF and travel to LA from Nov 2019 to Dec 2019. Our goal is to recommend LA restaurants for the sampled customers. 
For historical transactions, we use customers' SF transactions from May 2019 to Oct 2019. Utilizing the recommendation model for a customer, we obtain the top $15$ restaurants from the ranked list of LA restaurants and compute the score as a percentage of recommended restaurants in the real restaurant transactions made by the customer in LA from Nov 2019 to Dec 2019. 
For this test case, our original Word2Vec embeddings give a score of $60.36\%$.  
After removing the location feature between SF and LA, the score increases to $63.60\%$, while MUSE does not improve the score. 
%For experiment results with different $\alpha$, please refer Appendix~\ref{sec:rest-result}.

To test the effectiveness of UCAN under a large range of $\alpha$, we vary $\alpha$ from $0.5$ to $5$. All the alphas can work consistently well and $\alpha = 2$ gives the best results. Using $\alpha = 1$ as the rule of thumb can generally give good results.
Table~\ref{table_cross_alpha} shows that our algorithm outperforms the original Word2Vec embeddings on this task regardless of the value of $\alpha$ (beyond our default $\alpha=1$).

\begin{table}[]
\caption{Performance of the cross-city restaurant recommendations of UCAN with different $\alpha$.}
\centering
%\resizebox{0.95\textwidth}{!}{ % If your table exceeds the column or page width, use this command to reduce it slightly
\begin{tabular}{l||c}
\hline
& Score \\ 
\hline
Raw data  &  60.36  \\
MUSE & 60.33 \\
$\alpha=5$  &  63.51 \\
$\alpha=2$  &  64.11 \\
$\alpha=1$&  63.60\\
$\alpha=0.5$&  63.51 \\
\hline
\end{tabular}
\label{table_cross_alpha}
\end{table}
\section{Conclusion}
\label{sec:conclusion}
%We propose a general and effective method to measure and attenuate features of entities in an embedding. A domain adversarial network based algorithm is proposed to attenuate these features in the embeddings to aid in downstream tasks were these features are a nuisance.  We show that the proposed algorithm can significantly outperform the state-of-the-art algorithms on several datasets including graph neural network generated embedding and multi-language embedding.

We provide new and very general methods to measure and attenuate features from embeddings, and demonstrate effectiveness on four data sets and two novel downstream tasks.  Our key contribution is UCAN, an alignment algorithm which using the DAN framework to learn an unrestricted mapping from data with one feature label to data with the other, but adding a simple cosine similarity constraint to retain the structure.  We demonstrate that UCAN is a simple and effective method to refine embeddings.  

%In this paper, we propose a general and effective method to measure and attenuate features of entities in an embedding. This method is based on the Domain Adversarial Network (DAN), and considers all possible transformations, not just those restricted to affine steps (like projection, rotation, translation).  It ensures that the embedding does not lose meaning by penalizing large changes of points under the cosine distance.  Extensive experiments demonstrate that the proposed algorithm can significantly outperform state-of-the-art algorithms on multiple embedding datasets. We also show a downstream task on detecting nuisance features using embeddings generated by our method with attenuated features.
\section*{Acknowledgment}
We thank our support from NSF CCF-1350888, CNS-1514520, CNS-1564287, IIS-1816149, CCF-2115677, and from Visa Research.

\bibliographystyle{IEEEtran}
\bibliography{paper}

\end{document}